\documentclass[letterpaper, 10 pt, conference]{ieeeconf}  

\IEEEoverridecommandlockouts                              

\usepackage{amsmath} 
\usepackage{amssymb}  

\usepackage{graphicx}
\usepackage{subcaption}
\usepackage[export]{adjustbox}

\usepackage{tabularx}
\usepackage{multirow}

\usepackage{comment}

\usepackage{booktabs} 

\title{\LARGE \bf
FGO-SLAM: Enhancing Gaussian SLAM with Globally Consistent Opacity Radiance Field
}

\author{Fan Zhu$^{1,2}$, Yifan Zhao$^{3}$, Ziyu Chen$^{1,2}$, Biao Yu$^{1}$, and Hui Zhu$^{1}$
\thanks{$^{1}$ Hefei Institutes of Physical Science, Chinese Academy of Sciences, Hefei, China.}%
\thanks{$^{2}$ Department of Automation, University of Science and Technology of China, Hefei, China.}%
\thanks{$^{3}$ Department of Mathematics, University of Science and Technology of China, Hefei, China.}%
}

\begin{document}

\maketitle
\thispagestyle{empty}
\pagestyle{empty}

\begin{abstract}

Visual SLAM has regained attention due to its ability to provide perceptual capabilities and simulation test data for Embodied AI. However, traditional SLAM methods struggle to meet the demands of high-quality scene reconstruction, and Gaussian SLAM systems, despite their rapid rendering and high-quality mapping capabilities, lack effective pose optimization methods and face challenges in geometric reconstruction. To address these issues, we introduce FGO-SLAM, a Gaussian SLAM system that employs an opacity radiance field as the scene representation to enhance geometric mapping performance. After initial pose estimation, we apply global adjustment to optimize camera poses and sparse point cloud, ensuring robust tracking of our approach. Additionally, we maintain a globally consistent opacity radiance field based on 3D Gaussians and introduce depth distortion and normal consistency terms to refine the scene representation. Furthermore, after constructing tetrahedral grids, we identify level sets to directly extract surfaces from 3D Gaussians. Results across various real-world and large-scale synthetic datasets demonstrate that our method achieves state-of-the-art tracking accuracy and mapping performance.

\end{abstract}

\section{INTRODUCTION}

Embodied Artificial Intelligence (Embodied AI) has garnered significant attention as a critical component in Artificial General Intelligence (AGI). To achieve Embodied AI, dense Visual SLAM methods are required that not only provides embodied perception for real-world agents but also reconstructs high-quality, realistic scene maps in cyberspace for simulation testing \cite{EAI_OV}. Traditional Visual SLAM systems \cite{Orb-slam2, ORB3, SVO, ElasticFusion} employ explicit representations such as point clouds, voxels, and TSDF, achieving promising results in early research stages. However, the low-resolution maps with holes and lack of texture details generated by traditional dense Visual SLAM methods no longer meet the demands of Embodied AI, AR/VR, and similar applications. Therefore, a novel Visual SLAM method is needed to provide high quality scene datasets for simulation testing on the basis of satisfying the embodied perception.
 
The advent of Neural Radiance Fields (NeRF) \cite{Nerf} and its subsequent developments \cite{NGP} has introduced a novel approach to scene representation for robotics \cite{Nerf2real, Nerf-vins, Rapid-Mapping}. NeRF-based SLAM \cite{Imap, Orbeez-slam, Ngel-slam} has shown the potential for integrating neural implicit representations into SLAM systems to reconstruct high-quality scenes. However, the implicit representation method using multilayer perceptrons (MLP) may result in catastrophic forgetting and over-smoothing, which challenges the ability of NeRF-based dense Visual SLAM systems to achieve large-scale reconstruction and accurately capture geometric features.

\begin{figure}[!tpb]
	\centering
	\includegraphics[width=\columnwidth]{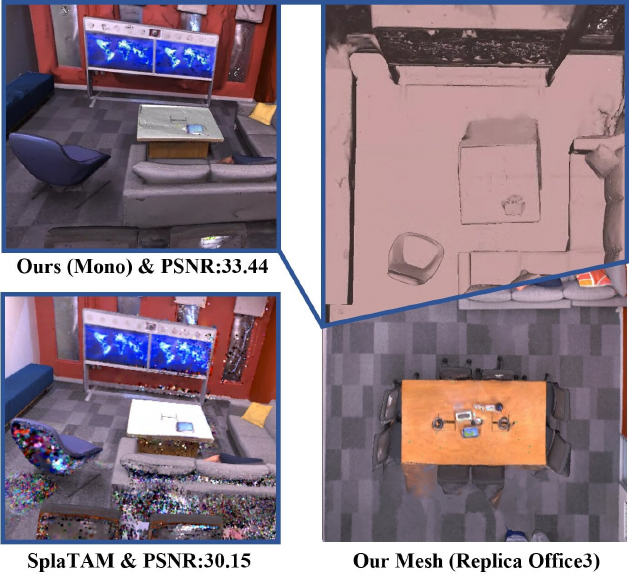}
	\caption{\textbf{Comparison of mapping effect.} Our method constructs a complete mesh and provides higher-fidelity images along with more accurate camera pose tracking results. This method does not rely on truth depth images for map optimization, which allows us to work with a monocular camera.}
	\label{fig:abstract}
\end{figure}

Recently, 3D Gaussian Splatting (3DGS) \cite{3dgs} has employed Gaussian primitives as an explicit representation, retaining NeRF's advantages in high-quality scene reconstruction while being easier to optimize and achieving faster rendering speeds. 3DGS-based SLAM methods achieve localization accuracy comparable to NeRF-based systems and generate higher-quality novel views. However, existing Gaussian SLAM systems lack effective camera pose optimization methods, restricting their application in complex environments. Additionally, the explicit and non-coherent nature of 3D Gaussian distributions presents challenges for surface reconstruction and mesh generation.

To tackle these challenges, we introduce a Gaussian SLAM method that employs a globally consistent opacity radiance field as its underlying representation. FGO-SLAM offers real-time, precise tracking capabilities for agents and provides high-quality geometric scene maps for simulation training in virtual environments, as shown in Fig. \ref{fig:abstract}. 
 
Specifically, we enhance tracking and mapping to maintain a globally consistent opacity radiance field, and provide surface extraction capabilities. We build our tracking module based on the traditional feature-based visual odometry (VO) approach \cite{ORB3}, which allows us to easily adjust global poses and point cloud positions, thereby achieving robust tracking. In mapping, we construct a 3D Gaussian-based opacity radiance field on sparse point clouds, using an explicit ray-Gaussian approach to determine the opacity of any point in 3D space, forming the basis for high-quality geometric reconstruction. Additionally, we incorporated two regularization terms: normal consistency and depth loss independent of priors, which further enhance the mapping performance of our method. For surface extraction, unlike Poisson reconstruction or TSDF fusion, we directly extract surfaces from 3D Gaussians by identifying level sets. This approach is enabled by our constructed opacity radiance field, which allows the formation of tetrahedral grids for each 3D Gaussian primitive. After evaluating the opacity of tetrahedral vertices, we identify level sets through binary search and perform mesh extraction using the marching tetrahedrons algorithm. In summary, our contributions are as follows:

\begin{itemize}
	\item{We introduce FGO-SLAM, a Gaussian SLAM method that uses a globally consistent opacity radiance field for mapping, capable of robustly tracking and providing high-quality scene datasets for agent simulation testing.}
	
	\item{We propose a globally consistent opacity radiance field for scene representation, incorporating depth distortion and normal consistency as regularization terms, enabling efficient scene geometry fitting and overcoming mesh extraction challenges in existing Gaussian SLAM systems.}
	
	\item{We performed extensive experiments on challenging real-world and large-scale synthetic datasets, showing that our method achieves state-of-the-art (SOTA) quantitative and qualitative results.}
\end{itemize}

\section{Related Works}

\subsection{Dense Visual SLAM}

Using visual SLAM for dense mapping is crucial for robotics, autonomous driving, and AR/VR. Early studies typically employed explicit representation methods such as point clouds and voxels for dense reconstruction \cite{3DR_SOTA}. DTAM \cite{Dtam} first introduced a dense SLAM system using photometric consistency for camera tracking and representing the scene as a cost volume. KinectFusion \cite{Kinectfusion} performs camera tracking through iterative closest point and updates the scene using TSDF Fusion. DI-Fusion \cite{Di-fusion} fits scene geometry with locally implicit fields modeled by deep networks, allowing incremental reconstruction. These traditional dense SLAM methods are mature in tracking but limited in providing high-fidelity scene models and generalized reasoning capabilities.

\subsection{NeRF-based Visual SLAM}

In recent years, dense NeRF-based Visual SLAM \cite{Imap, Nice-slam, HEROSLAM} have gained significant attention. iMAP \cite{Imap} first demonstrated the potential of neural implicit representation in SLAM. However, a single MLP often loses tracking in large-scale scenes. NICE-SLAM \cite{Nice-slam} achieved good results in large scenes using pretrained parameters and hierarchical feature grids, but suffers from drift issues. ESLAM \cite{Eslam} represents scenes with multi-scale axis-aligned feature planes, and Co-SLAM \cite{Co-slam} uses multi-resolution hash-grids, both showing impressive mapping capabilities. However, these NeRF-based methods typically lack loop closing correction. Additionally, the implicit representation of MLP can lead to catastrophic loss and over-smoothing, making it challenging for NeRF-based dense visual SLAM systems to perform large-scale reconstruction and capture precise geometric features.

\subsection{3DGS-based Visual SLAM}
Recently, 3DGS \cite{3dgs} has emerged as a significant research direction in computer vision\cite{2dgs} and robotics \cite{Liv-gaussmap, Photo-slam}, offering a more efficient paradigm for scene representation and reconstruction. GS-SLAM \cite{Gs-slam} utilizes three-dimensional Gaussian distributions, opacity, and spherical harmonics to encapsulate scene geometry and appearance, enhancing map optimization and rendering speed. Photo-SLAM \cite{Photo-slam} introduces a Gaussian pyramid training method to achieve photorealistic image synthesis. SplaTAM \cite{Splatam} employs isotropic Gaussians and optimization guided by differentiable rendering silhouettes for accurate and efficient camera tracking and map construction. MonoGS \cite{MonoGS} uses 3D Gaussians as the sole map representation, enabling real-time scene capture. However, current Gaussian SLAM systems lack effective camera pose optimization methods, limiting their application in complex scenes. Furthermore, the explicit and disconnected nature of 3D Gaussian distributions poses challenges for surface reconstruction and mesh generation.

\section{Method}

\begin{figure*}[!htpb]
	\centering
	\includegraphics[width=\textwidth]{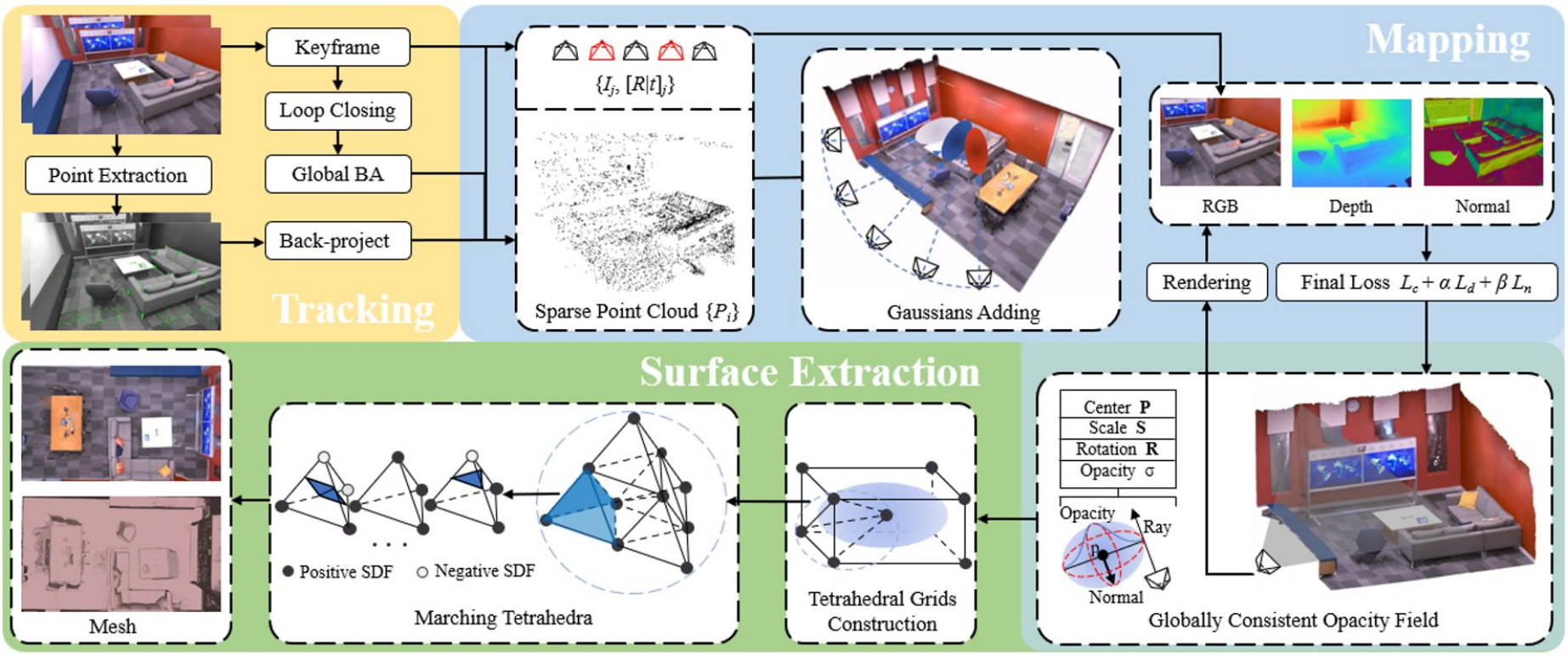}
	\caption{\textbf{Overview.} Our proposed system is composed of three modules divided into two stages: the tracking and mapping modules in the first stage, and the mesh extraction module in the second stage. The tracking and mapping modules maintain a globally consistent opacity radiance field as the foundational map. FGO-SLAM accepts RGB or RGB-D image frames, where the tracking module estimates the camera pose in real time. Upon detecting a loop closing, the system corrects the poses of all keyframes and point cloud positions, producing an image sequence with accurate poses and sparse point clouds for the mapping module. The mapping module incrementally constructs the Gaussian underlying map using the opacity radiance field, continuously optimizing the overall structure. Finally, the Surface Extraction module constructs bounding boxes around each Gaussian primitive and employs the Marching Tetrahedra algorithm to extract geometric structures.}
	\label{fig:Framework}
\end{figure*}

The system pipeline of FGO-SLAM is shown in Fig. \ref{fig:Framework}. Unlike other 3DGS-based SLAM systems, our approach incorporates a Surface Extraction module in addition to the conventional tracking and mapping modules, directly using Gaussian primitives to construct meshes, thanks to our opacity radiance field map representation. This section introduces our system from the following perspectives: Sec. \ref{sec:tracking} describes the tracking module with global optimization. Sec. \ref{sec:ORF} details the opacity radiance field map representation. Sec. \ref{sec:opt} covers the map optimization methods. Sec. \ref{sec:surface} describes the surface extraction method.

\subsection{Tracking with Global Adjustment}
\label{sec:tracking}

\subsubsection{Pose Estimation}

Our tracking module estimates 6-DoF camera poses and generates a sparse set of 3D point clouds from the input images. Similar to \cite{ORB3}, geometric feature points are extracted from consecutive frames, and camera poses are estimated using the reprojection error:

\begin{equation}
\mathcal{L}_P=\sum_{i,j}\rho\left(\left\|\mathbf{p}_i-\pi\left(\mathbf{R}_j\mathbf{P}_i+\mathbf{t}_j\right)\right\|_{\Sigma_g}^2\right)
\end{equation}

Here, $\mathbf{R}_j\in SO^3$ and $\mathbf{t}_j\in\mathbb{R}^3$ represent the orientation and position of the j-th camera, $\pi\begin{pmatrix}\cdot\end{pmatrix}$ denotes the projection of the 3D point $\mathbf{P}_i\in\mathbb{R}^3$ into 2D space, $\mathbf{p}_i$ is the corresponding 2D matched point, $i\in N$ is the index of the matching set, $\mathbf{\Sigma}_{g}$ is the scale-associated covariance matrix of the keypoint, and $\rho$ denotes the robust Huber cost function. We optimize the camera pose by minimizing the reprojection error:

\begin{equation}
\{\mathbf{R}_j,\mathbf{t}_j\}=\underset{R_j,t_j}{\operatorname*{\arg\min}}\mathcal{L}_P
\end{equation}

\subsubsection{Global Adjustment}

The global adjustment relies on the loop closing, which is a vital component in SLAM systems, correcting accumulated tracking errors and drift to maintain global consistency. We employ a Bag of Words model \cite{BoW} for loop keyframe recognition and similarity verification. Upon meeting loop closing conditions, we correct local keyframe poses using similarity transformations and refine the system's keyframes and point clouds via global Bundle Adjustment (BA). 

We optimize the reprojection error between 3D points $\mathbf{P}_{i}$ and camera poses $\{\mathbf{R}_j,\mathbf{t}_j\}$:

\begin{equation}
\{\mathbf{R}_j,\mathbf{t}_j,\mathbf{P}_i\}=\arg\min_{\mathbf{R}_j,\mathbf{t}_j,\mathbf{P}_i}\mathcal{L}_P
\end{equation}

And we solve this nonlinear least squares problem using the Levenberg-Marquardt method. Global adjustment effectively eliminates ghosting from odometry drift, improving overall mapping quality.

\subsection{Opacity Radiance Field}
\label{sec:ORF}

Following 3DGS \cite{3dgs}, we build the underlying scene using a set of 3D Gaussian primitives. Each Gaussian $\mathcal{G}=\delta N(\mathbf{\mu},\mathbf{\Sigma})$ is defined by its opacity $\delta\in\begin{bmatrix}0,1\end{bmatrix}$, mean $\mathbf{\mu}\in\mathbb{R}^{3\times1}$ in world coordinates, covariance matrix $\mathbf{\Sigma}\in\mathbb{R}^{3\times3}$ formed by scale $\mathbf{S}\in\mathbb{R}^{3\times3}$ and rotation $\mathbf{R}\in\mathbb{R}^{3\times1}$, and an associated color parameter $\mathbf{c}\in\mathbb{R}^{3\times1}$. 

However, the explicit and disconnected nature of 3D Gaussian primitives makes it difficult for 3DGS-based SLAM systems to perform surface reconstruction. Inspired by \cite{gof}, we evaluate each 3D Gaussian through ray-Gaussian intersection, assessing it as a 1D Gaussian function rather than projecting it onto a 2D plane as a 2D Gaussian. In the local coordinate system of a 3D Gaussian, along a ray, the Gaussian value at a point $\mathbf{x}_g\in\mathbb{R}^{3\times1}$ at depth $d$ is: 

\begin{equation}
	\mathcal{G}(\mathbf{x}_g)=\exp(-\frac12\mathbf{x}_g^T\mathbf{x}_g)
	\label{eq:Gxg}
\end{equation}

Here, $\mathbf{x}_g=\mathbf{o}_g+d\mathbf{r}_g$, where $\mathbf{o}_g\in\mathbb{R}^{3\times3}$ denotes the camera center, $\mathbf{r}_g\in\mathbb{R}^{3\times1}$ is the direction of the ray, and the subscript $g$ refers to the Gaussian's local coordinate system. Substituting $\mathbf{x}_g$ into the \eqref{eq:Gxg}, we derive the 1D Gaussian value at any point along the ray within the local coordinate system:

\begin{equation}
	\mathcal{G}^{1D}(d)=\exp\biggl(-\frac12\biggl(\mathbf{o}_g+d\mathbf{r}_g\biggr)^T\biggl(\mathbf{o}_g+d\mathbf{r}_g\biggr)\biggr)
\end{equation}

Clearly, when $d=-\frac{\mathbf{r}_g^T\mathbf{r}_g}{\mathbf{0}_g^T\mathbf{r}_g}$ is reached, the function $\mathcal{G}^{1D}(d)$ reaches its maximum, denoted as $\mathcal{G}_{\max}^{1D}$. This means that if there is only one Gaussian $\mathcal{G}$ along the ray, the opacity increases along the ray until it reaches $\mathcal{G}_{\max}^{1D}$, after which it remains constant. Thus, we construct the opacity radiance field using each Gaussian's opacity, defining the opacity of any 3D point $\mathbf{P}$ as the minimum opacity value across all views: 

\begin{equation}
	\mathcal{O}(\mathbf{P})=\min_{(r,t)}\mathcal{O}(\mathbf{0},\mathbf{r},d)
\end{equation}

Here, $\mathcal{O}(\mathbf{0},\mathbf{r},d)$ represents the opacity value for a single view, similar to volume rendering:

\begin{equation}
	\mathcal{O}(\mathbf{o},\mathbf{r},d)=\sum_{i\in N}\delta_i\mathcal{G}_i^{1D}(d)\prod_{j=1}^{i-1}\left(1-\delta_j\mathcal{G}_i^{1D}(d)\right)
\end{equation}

\subsection{Scene Optimization}
\label{sec:opt}

\begin{figure}[!tpb]
	\centering
	\includegraphics[width=\columnwidth]{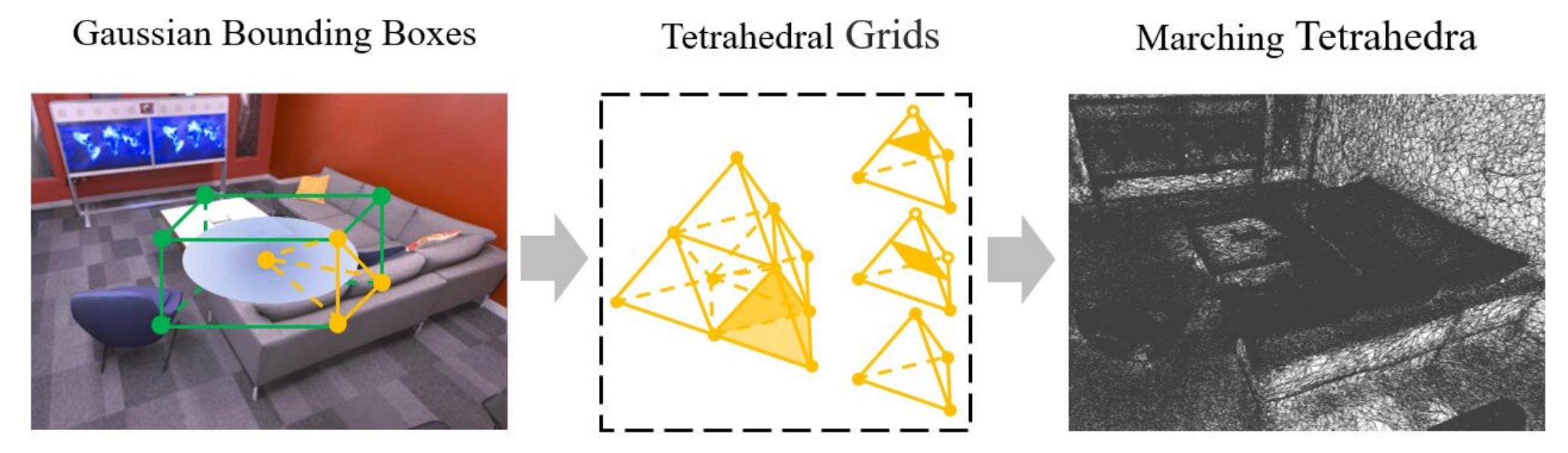}
	\caption{\textbf{Surface Extraction.}}
	\label{fig:surface_ex}
\end{figure}

\begin{table*}[!htpb]
	\caption{\textbf{Mapping Results on Replica \cite{replica}.} Data averaged from eight scenes.}
	\centering
	\begin{tabularx}{\textwidth}{lccccccc}
			\toprule
			\textbf{}  & NICE-SLAM \cite{Nice-slam} & ESLAM \cite{Eslam} & Photo-SLAM \cite{Photo-slam} & SplaTAM \cite{Splatam} & MonoGS \cite{MonoGS} & \textbf{Ours (Mono)} & \textbf{Ours (RGB-D)}          \\
			\midrule
			PSNR ↑      & 25.90     & 29.92   & 34.96  & 34.20   & 36.75  & 34.13      & \textbf{38.35} \\
			SSIM ↑      & 0.822     & 0.896   & 0.942  & 0.970   & 0.960  & 0.956      & \textbf{0.973} \\
			LPIPS ↓     & 0.238     & 0.217   & \textbf{0.059}  & 0.099   & 0.085  & 0.094      & 0.084 \\
			Depth L1 [cm] ↓ & 3.53      & 1.18    & -  & 0.73       & \textbf{0.69}      & 3.24          & 0.71              \\
			\bottomrule
		\end{tabularx}
	\label{tab:replica}
\end{table*}

To broaden the application scope of our system, we aim to avoid relying on depth priors for mapping. However, 3D reconstruction with monocular SLAM is inherently under-constrained, and optimizing based only on color loss tends to introduce noise. Inspired by \cite{2dgs, gof}, we expanded the regularization terms to further optimize our Gaussian map. We extract the parameter $\omega_i=\delta_i\mathcal{G}_{i,\max}^{1D}\prod_{j=1}^{i-1}\left(1-\delta_j\mathcal{G}_{i,\max}^{1D}\right)$, which represents the mixing weight of the i-th Gaussian $\mathcal{G}$.

\subsubsection{Color Loss}
We employ a combination of L1 loss for the photometric error between rendered and original images, along with the D-SSIM term, as the RGB reconstruction loss:

\begin{equation}
	\mathcal{L}_c=(1-\lambda)\mathcal{L}_1+\lambda\mathcal{L}_{D-SSIM}
\end{equation}

Unlike \cite{3dgs}, we determine the color of camera rays through alpha blending, rendered according to the original depth order:

\begin{equation}
	\mathcal{C}(\mathbf{o},\mathbf{r})=\sum_{i\in N}\omega_i\mathbf{c}_i
\end{equation}

\subsubsection{Depth Distortion}
We apply a depth distortion loss to align the Gaussian functions along the ray:

\begin{equation}
	\mathcal{L}_d=\sum_{i,j}\omega_i\omega_j\left|d_i-d_j\right|
\end{equation}

Here, $d_i$ and $d_j$ denote the distances at which the corresponding Gaussian function $\mathcal{G}_i^{1D}(d)$ attains its maximum value. To prevent exaggerated floating effects from prematurely blended Gaussians, we decouple the gradients of weights $\omega_i$ and $\omega_j$, focusing solely on minimizing the distance $d$. In this way, we get rid of the need for deep truth images, which enables our method to have a wider range of applications.

\subsubsection{Normal Consistency}
We define the normal of a 3D Gaussian as the normal of the intersection plane between the ray and the Gaussian, applying depth-normal consistency regularization: 

\begin{equation}
	\mathcal{L}_n=\sum_i\omega_i\left(1-\mathbf{n}_i^T\mathbf{N}\right)
\end{equation}

Here, $\mathrm{N}$ represents the normal estimated from the gradient of the depth map.

\subsubsection{Final Loss}
The final loss function is defined as:

\begin{equation}
	\mathcal{L}=\mathcal{L}_c+\alpha\mathcal{L}_d+\beta\mathcal{L}_n
\end{equation}

Where $\alpha$ ranges from 100 to 1000, and $\beta$ is set to 0.05.

\subsection{Surface Extraction}
\begin{figure}[!tbp]
	\centering
	\begin{subfigure}[b]{0.24\columnwidth}
		\centering
		\includegraphics[width=\textwidth]{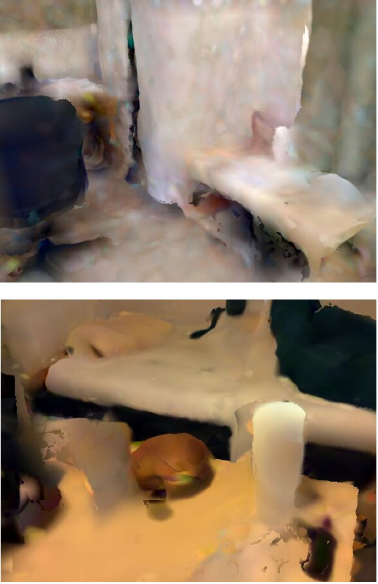}
		\caption{\footnotesize NICE-SLAM}
	\end{subfigure}
	\hfill
	\begin{subfigure}[b]{0.24\columnwidth}
		\centering
		\includegraphics[width=\textwidth]{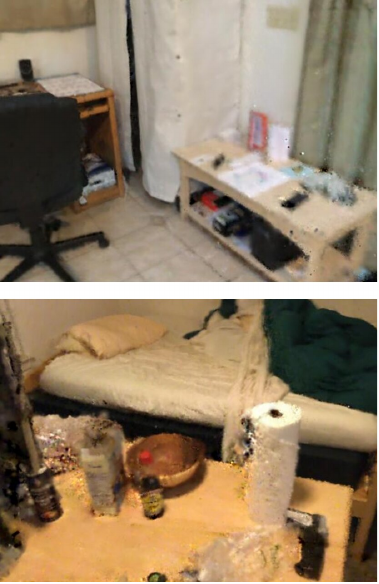}
		\caption{\footnotesize SplaTAM}
	\end{subfigure}
	\hfill
	\begin{subfigure}[b]{0.24\columnwidth}
		\centering
		\includegraphics[width=\textwidth]{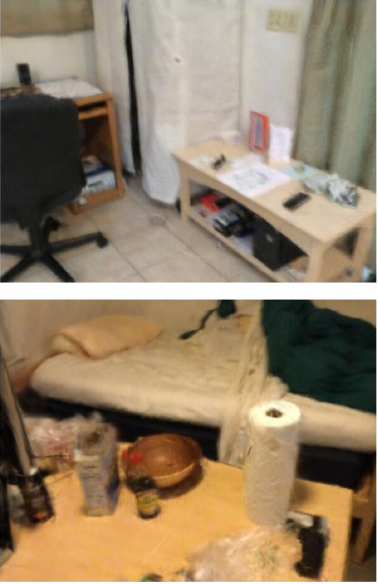}
		\caption{\footnotesize \textbf{Ours}}
	\end{subfigure}
	\hfill
	\begin{subfigure}[b]{0.24\columnwidth}
		\centering
		\includegraphics[width=\textwidth]{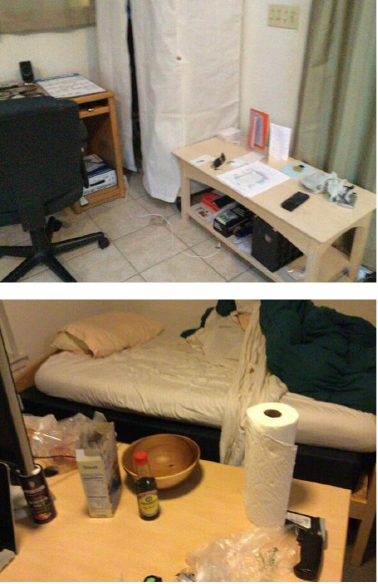}
		\caption{\footnotesize GT}
	\end{subfigure}
	\caption{\textbf{Visual comparison of the rendering quality on the ScanNet dataset \cite{Scannet}.} }
	\label{fig:render}
\end{figure}

Leveraging the opacity radiance field, we bypass Poisson reconstruction and TSDF fusion, directly extracting geometric shapes from 3D Gaussians using the Marching Tetrahedra algorithm \cite{Dmt}. As illustrated in Fig. \ref{fig:surface_ex}, we construct bounding boxes at the $3-\sigma$ level for each 3D Gaussian, generating tetrahedral grids from the box centers and vertices. Since the generated tetrahedral grids may connect distant points, leading to unreasonable geometric structures, we apply filtering. Specifically, when the length of an edge connecting two Gaussian primitives exceeds the sum of their maximum scales, the corresponding tetrahedral cell is removed. This ensures that the final tetrahedral units only contain edges connecting reasonable Gaussian distributions.

We follow the opacity evaluation method in \cite{gof} to evaluate the opacity of the tetrahedral grid, but different from it, we select key frames instead of all training views for evaluation, which reduces computational redundancy. Subsequently, we apply the Marching Tetrahedra algorithm to generate the initial mesh and identify the level sets by binary search.

\section{Experimental}

\subsection{Experimental Setup}

\subsubsection{Datasets}
We utilized three datasets to assess system performance: the classic real-world dataset TUM RGB-D \cite{tum}, the challenging ScanNet dataset \cite{Scannet} for tracking evaluation, and the synthetic Replica dataset \cite{replica} for mapping evaluation. The selection of scene sequences follows \cite{Nice-slam}.

\begin{table}[!tpb]
	\caption{\textbf{Tracking Results on TUM RGB-D \cite {tum}.} ATE-RMSE [cm] ($\downarrow$) is used as the metric.}
	\centering
	\begin{tabularx}{\columnwidth}{l *{4}{>{\centering\arraybackslash}X}}
		\toprule  %
		\textbf{}       & fr1/desk      & fr2/xyz       & fr3/office    & Avg.          \\ 
		\midrule
		NICE-SLAM \cite{Nice-slam}       & 2.70          & 1.80          & 3.00          & 2.50          \\
		ESLAM \cite{Eslam}          & 2.49          & 1.11          & 2.74          & 2.11          \\
		Photo-SLAM \cite{Photo-slam}     & 2.60          & \textbf{0.35}          & 1.00          & 1.31      \\
		SplaTAM \cite{Splatam}         & 3.34          & 1.34          & 5.21          & 3.30          \\
		MonoGS \cite{MonoGS}          & 1.52          & 1.58          & 1.65          & 1.58          \\
		\textbf{Ours (Mono)} 		       & 1.88          & 0.68          & 2.11          & 1.56          \\
		\textbf{Ours (RGB-D)}            & \textbf{1.49} & 0.48 & \textbf{0.98} & \textbf{0.98} \\ \bottomrule  
		\label{tab:tum}
	\end{tabularx}
\end{table}

\subsubsection{Baselines}
We compared our system's tracking and mapping performance against SOTA methods, including outstanding NeRF-based methods like NICE-SLAM \cite{Nice-slam} and ESLAM \cite{Eslam}, and recent 3D Gaussian-based methods such as Photo-SLAM \cite{Photo-slam}, SplaTAM \cite{Splatam}, and MonoGS \cite{MonoGS}. For some of the data, we used the results presented in their reports.

\subsubsection{Metrics}
We use root mean square error (RMSE) of absolute trajectory error (ATE) to evaluate tracking accuracy. For reconstruction quality, we use PSNR, SSIM, and LPIPS for rendered color images, and l1 loss for rendered depth images. For SLAM systems, real-time processing capability is important, as measured by frames per second (FPS). Based on \cite{Orbeez-slam}, we believe that quantitatively evaluating meshes performance may not be entirely fair, so we opted for a qualitative evaluation.

\subsubsection{Experiment Details}
We implemented FGO-SLAM on a computer with an Intel i7-12700K and an NVIDIA RTX 3090 Ti (24 GB). The sequences of these datasets are selected following NICE-SLAM \cite{Nice-slam}, with regularization parameters set to $\alpha = 1000 $ and $\beta = 0.05$. For a fair comparison with other methods, we also use RGB-D mode for our experiments. 

\begin{table}[!tpb]
	\caption{\textbf{Tracking Results on ScanNet \cite{Scannet}.} ATE-RMSE [cm] ($\downarrow$) is used as the metric.}
	\centering
		\begin{tabularx}{\columnwidth}{l *{7}{>{\centering\arraybackslash}X}}
			\toprule  
			\textbf{}  & 0000          & 0059          & 0106          & 0169          & 0181          & 0207          & Avg.           \\ 
			\midrule 
			NICE-SLAM \cite{Nice-slam}  & 9.76          & 12.83         & 8.03          & 10.49         & 13.09         & \textbf{5.79} & 10.00         \\
			ESLAM \cite{Eslam}      & 7.94          & 8.67          & 7.53          & 6.73          & \textbf{8.24} & 5.96          & 7.51          \\
			SplaTAM \cite{Splatam}    & 12.57         & 10.15         & 17.78         & 12.65         & 10.27         & 7.98          & 11.90         \\
			MonoGS \cite{MonoGS}     & 15.72         & 12.85         & 20.21         & 10.98         & 10.23         & 11.05         & 13.51         \\
			\textbf{Ours (Mono)}     & 7.25         & 7.09         & 8.02         & 7.15         & 12.32         & 8.26         & 8.35        \\
			\textbf{Ours (RGB-D)}       & \textbf{6.74} & \textbf{6.49} & \textbf{7.25} & \textbf{6.43} & 10.11         & 7.19          & \textbf{7.37} \\
			\bottomrule  
			\label{tab:scannet}
		\end{tabularx}
\end{table}

\subsection{Experiment Results}
\subsubsection{Evaluation on TUM RGB-D \cite{tum}}

We initially evaluated our method's camera tracking performance on the classic TUM RGB-D dataset. As shown in TABLE \ref{tab:tum}, our approach clearly outperforms the other baselines. The results demonstrate that our use of the conventional VO module and global adjustment has significant advantages in camera tracking.

\subsubsection{Evaluation on ScanNet \cite{Scannet}}

We evaluated our method's camera tracking accuracy on the larger and more challenging ScanNet dataset. TABLE \ref{tab:scannet} presents our tracking performance. Our method still maintains the leading tracking performance.

\label{sec:surface}
\begin{figure}[!tbp]
	\centering
	\begin{subfigure}[b]{0.24\columnwidth}
		\centering
		\includegraphics[width=\textwidth]{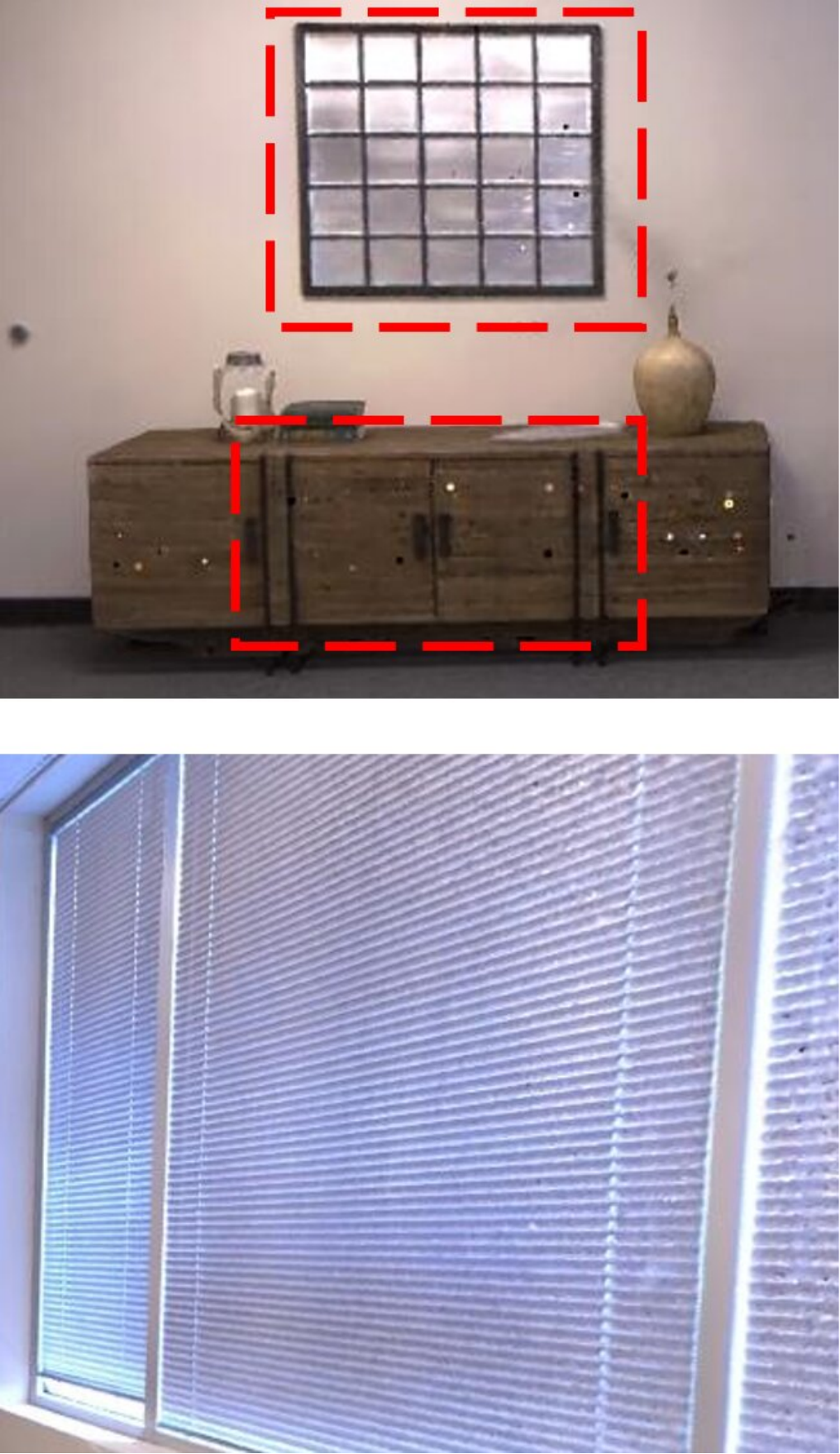}
		\caption{SplaTAM}
	\end{subfigure}
	\hfill
	\begin{subfigure}[b]{0.24\columnwidth}
		\centering
		\includegraphics[width=\textwidth]{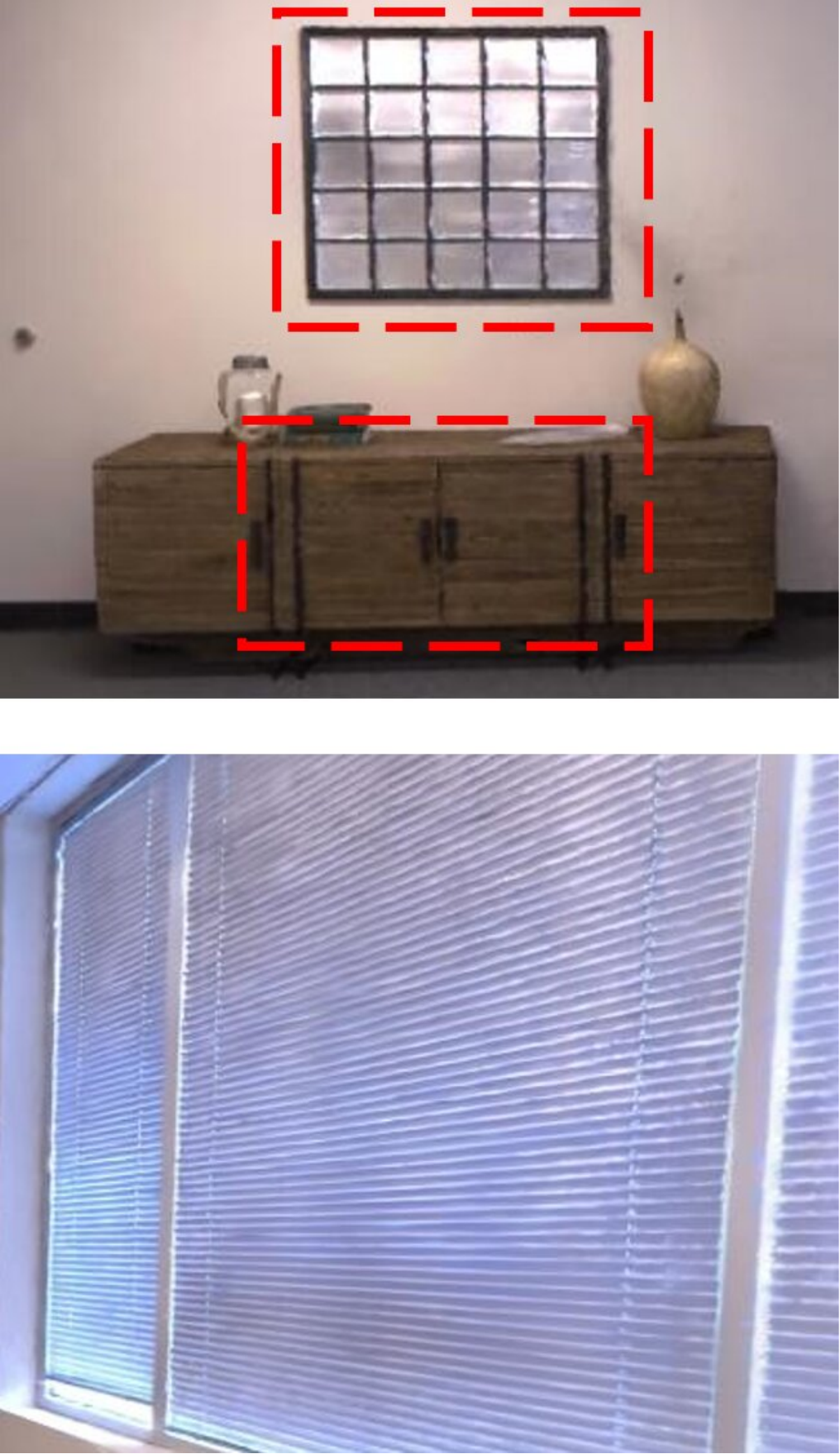}
		\caption{MonoGS}
	\end{subfigure}
	\hfill
	\begin{subfigure}[b]{0.24\columnwidth}
		\centering
		\includegraphics[width=\textwidth]{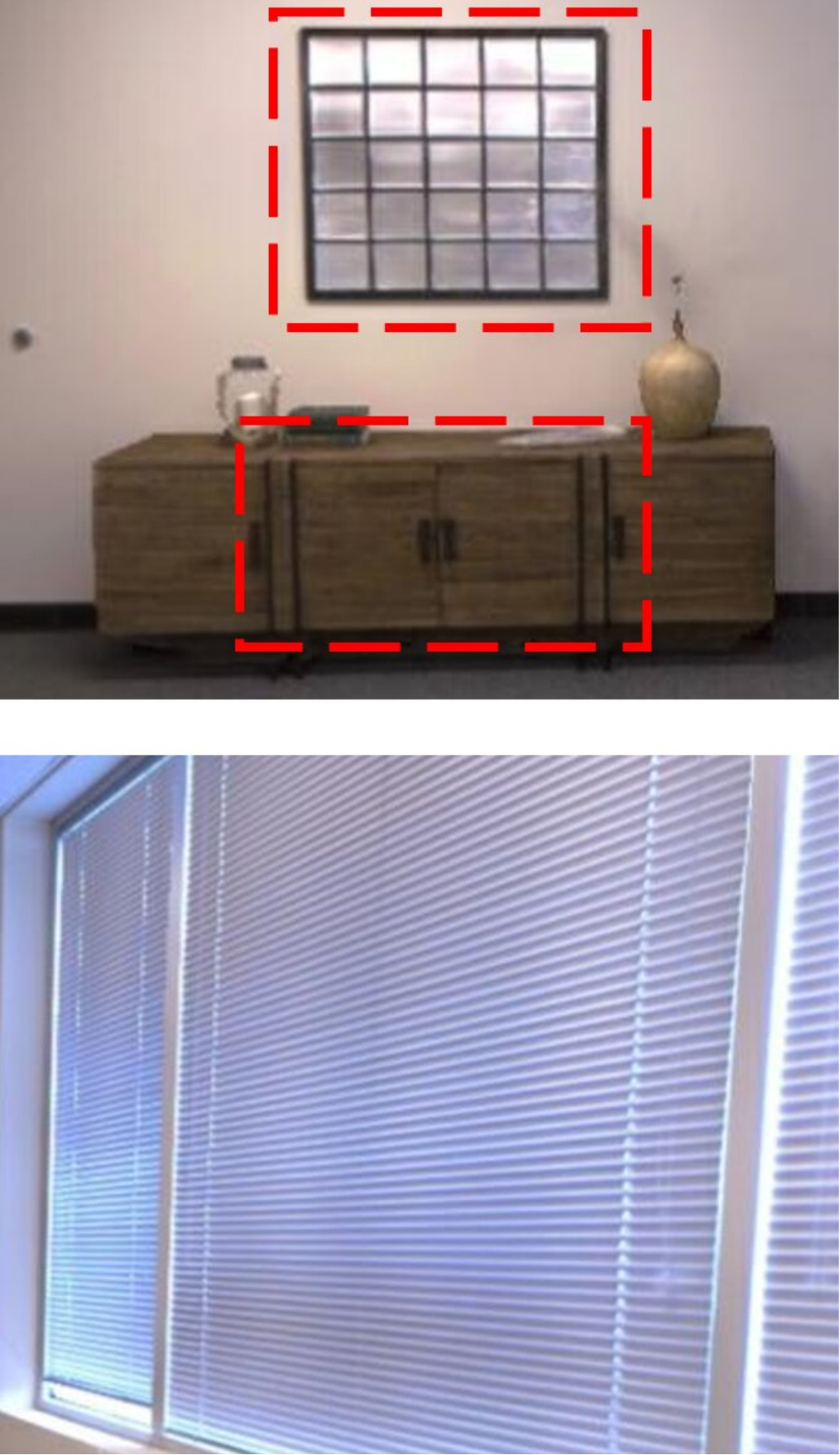}
		\caption{\textbf{Ours}}
	\end{subfigure}
	\hfill
	\begin{subfigure}[b]{0.24\columnwidth}
		\centering
		\includegraphics[width=\textwidth]{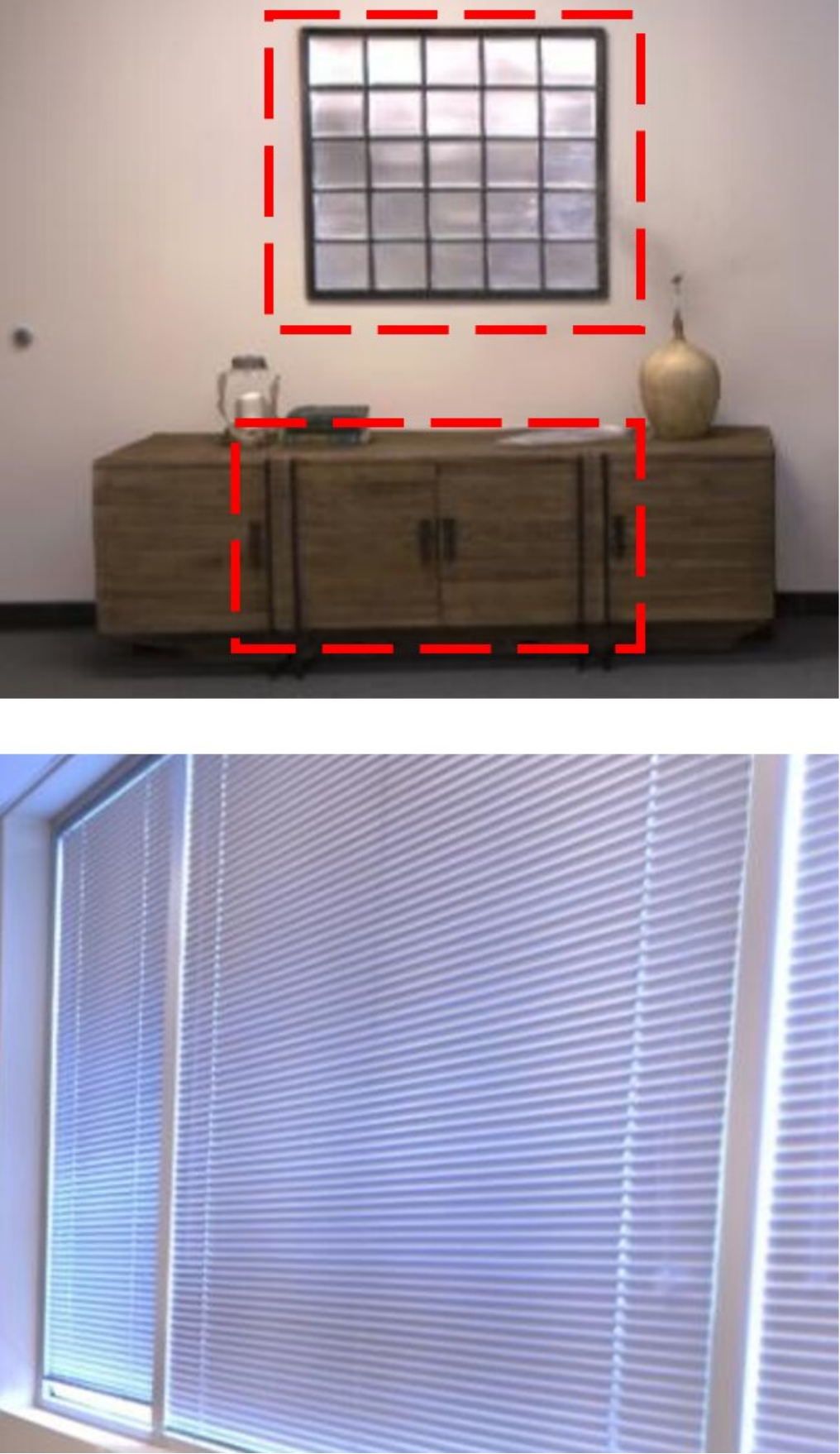}
		\caption{GT}
	\end{subfigure}
	\caption{\textbf{Rendering results on the Replica dataset \cite{replica}.} The dashed boxes highlight the differences between our approach and baselines. For the details of Windows and cabinets, our method significantly outperforms baselines.}
	\label{fig:replica}
\end{figure}

Additionally, Fig. \ref{fig:render} compares the rendering results of our method and the baseline methods on the ScanNet dataset. We can clearly notice that our method generates sharper images than those produced by NICEC-SLAM and SplaTAM.

\subsubsection{Evaluation on Replica \cite{replica}}

We assessed our method's scene reconstruction capabilities using the high-quality synthetic dataset, Replica. As indicated in TABLE \ref{tab:replica}, our approach achieves the best average rendering quality across all scenes in terms of rendering quality. Fig. \ref{fig:replica} provides a detailed comparison of our method against baseline methods, demonstrating that our rendered images are closer to the ground truth (GT) images in detail than the baseline methods. 

\begin{figure}[!tbp]
	\centering
	\begin{subfigure}[b]{0.24\columnwidth}
		\centering
		\includegraphics[width=\textwidth]{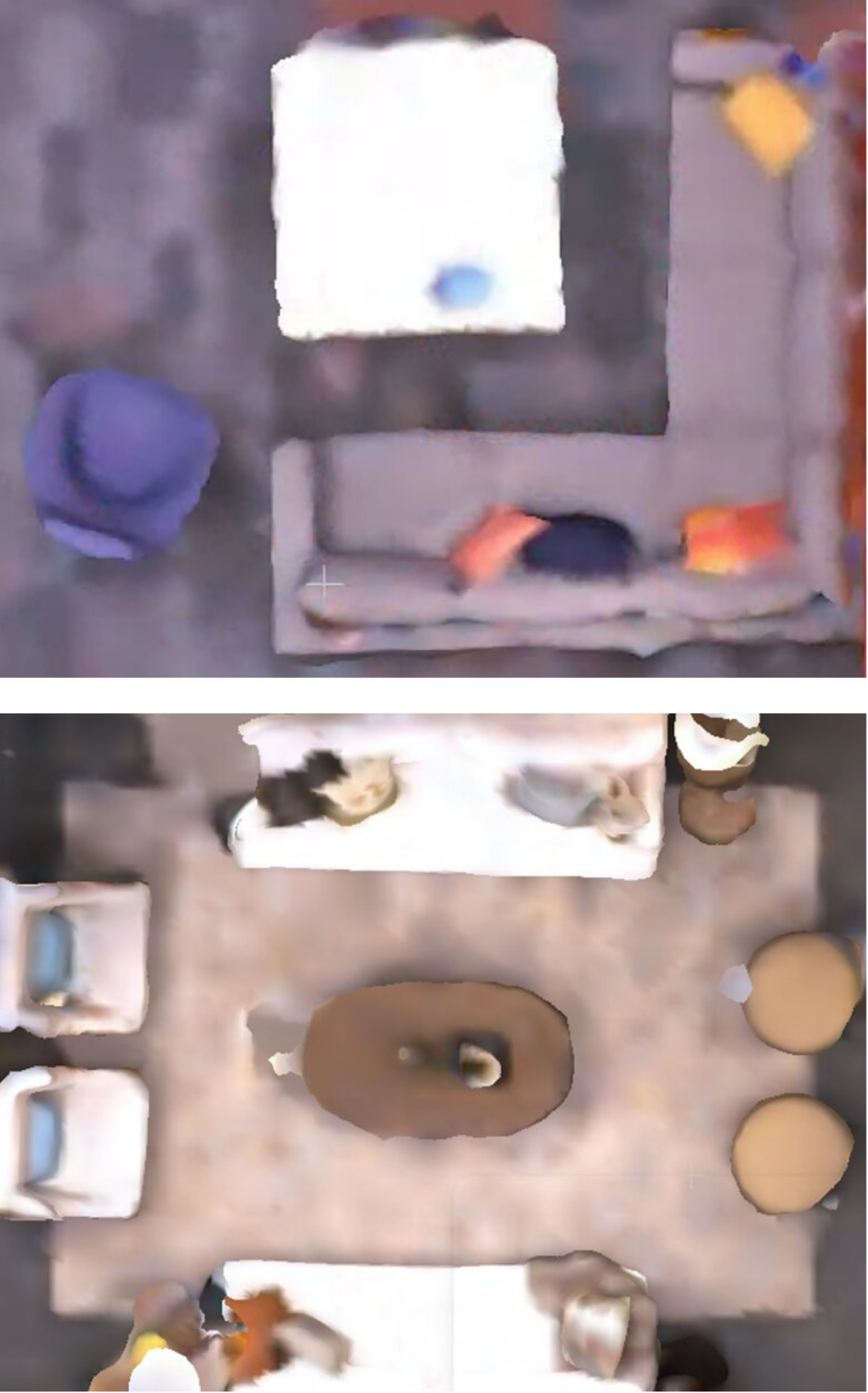}
		\caption{\footnotesize NICE-SLAM}
	\end{subfigure}
	\hfill
	\begin{subfigure}[b]{0.24\columnwidth}
		\centering
		\includegraphics[width=\textwidth]{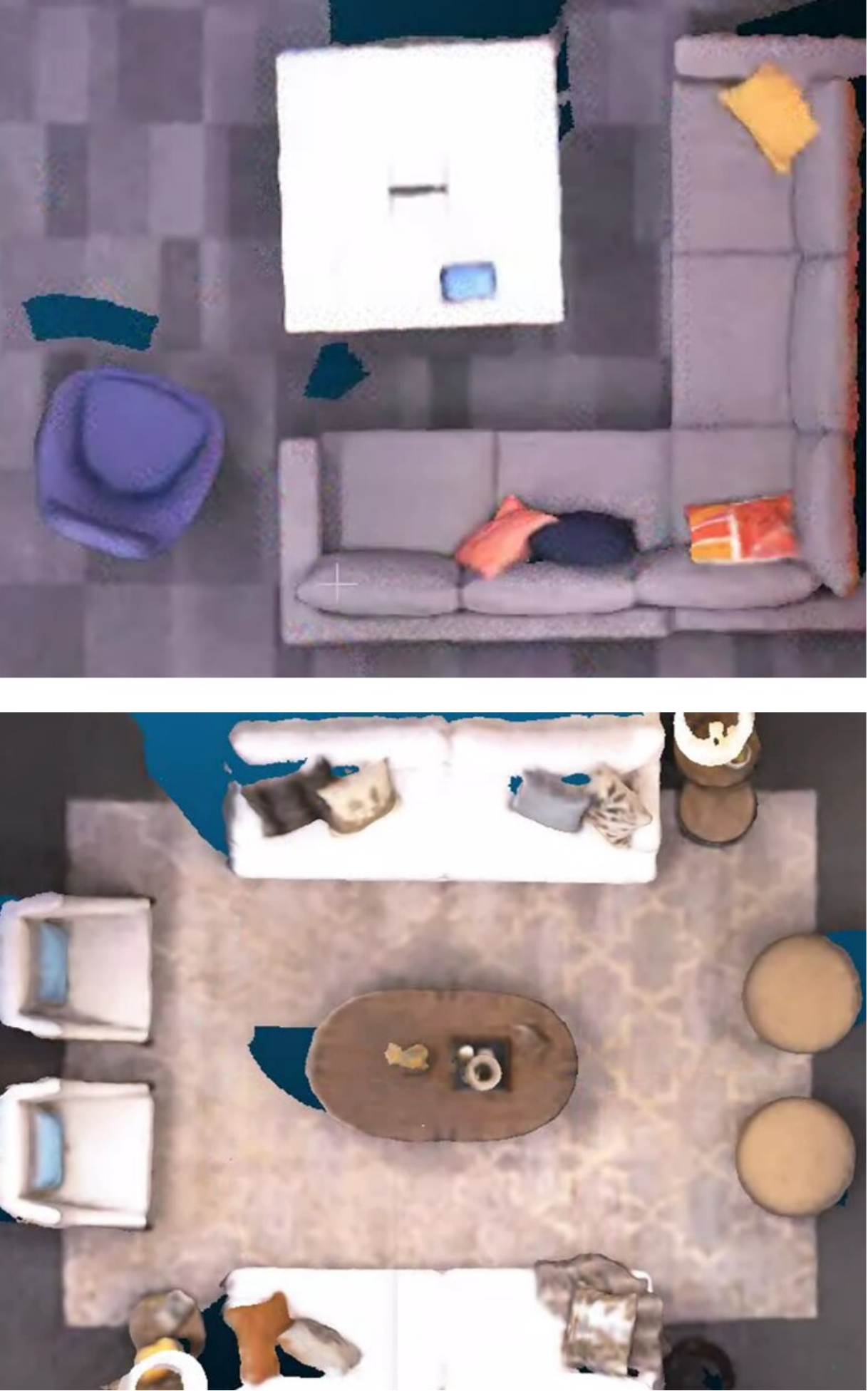}
		\caption{\footnotesize ESLAM}
	\end{subfigure}
	\hfill
	\begin{subfigure}[b]{0.24\columnwidth}
		\centering
		\includegraphics[width=\textwidth]{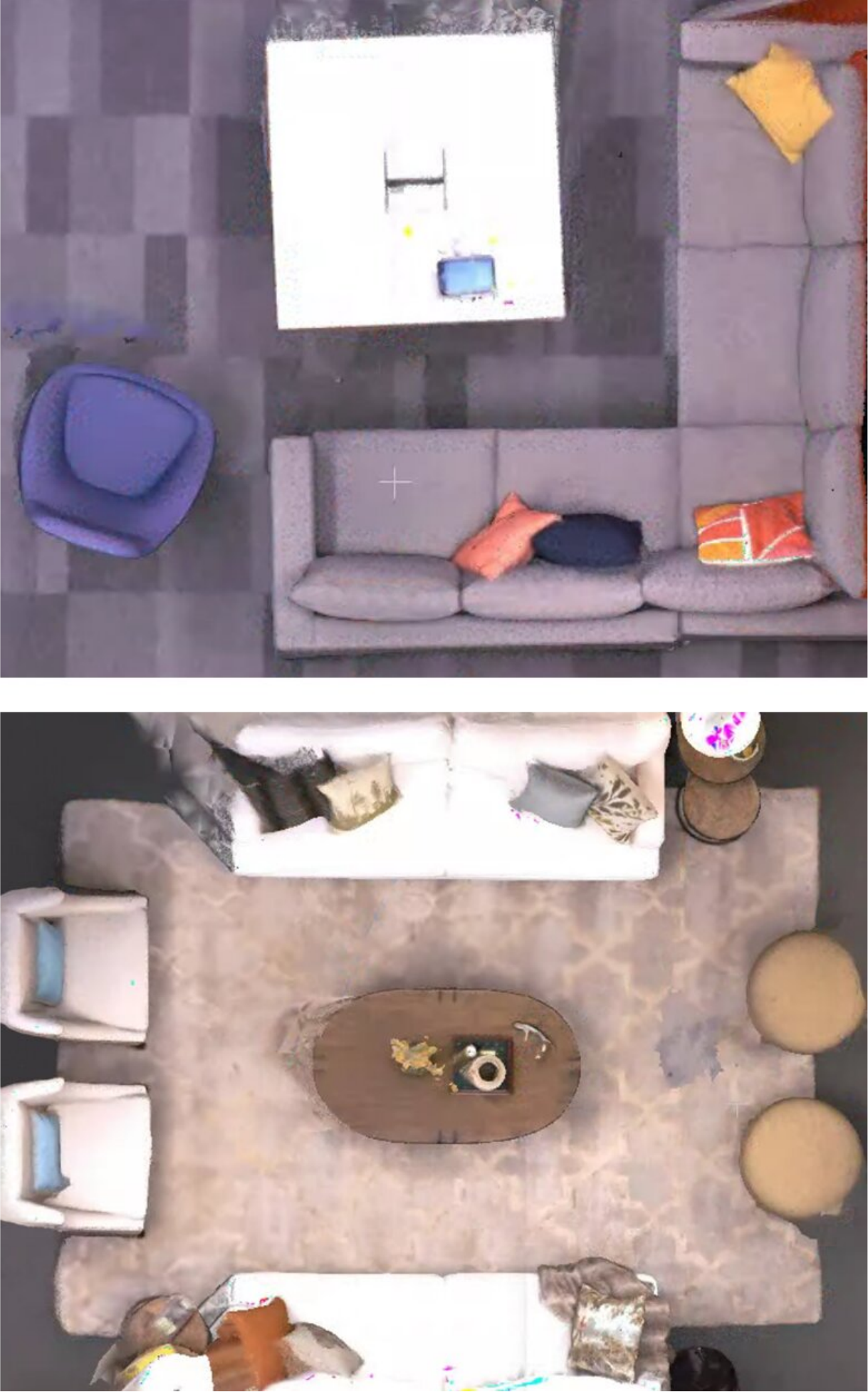}
		\caption{\footnotesize \textbf{Ours}}
	\end{subfigure}
	\hfill
	\begin{subfigure}[b]{0.24\columnwidth}
		\centering
		\includegraphics[width=\textwidth]{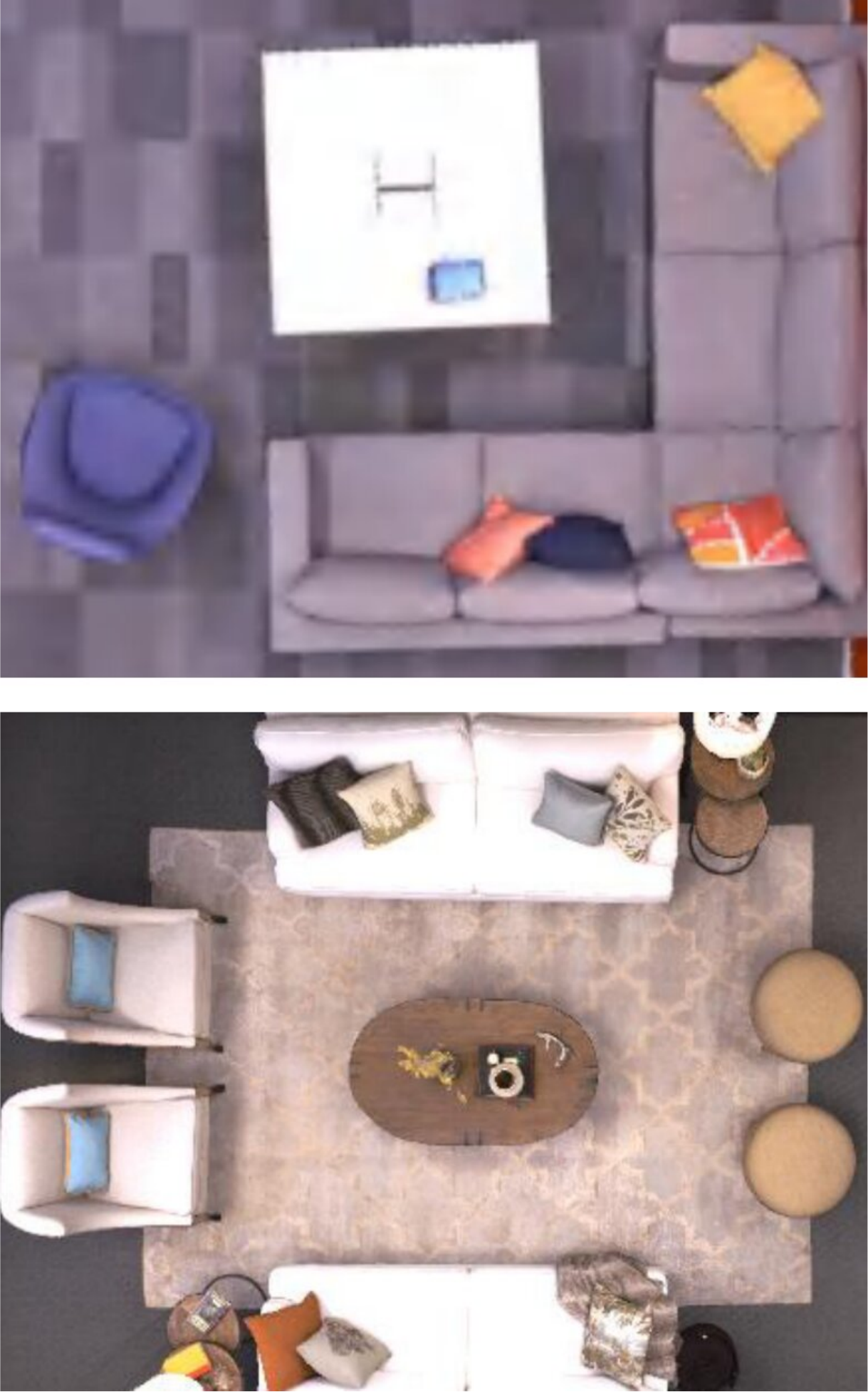}
		\caption{\footnotesize GT}
	\end{subfigure}
	\caption{\textbf{Visual comparison of the mesh quality on the Replica dataset \cite{replica}.} Previous Gaussian-based methods cannot generate mesh effectively, and therefore, cannot be compared here.}
	\label{fig:mesh}
\end{figure}

In terms of geometric map construction, our method has achieved results second only to MonoGS at Depth L1. Additionally, our approach efficiently generates meshes, a capability that other 3DGS-based SLAM methods lack, as shown in Fig. \ref{fig:abstract} and \ref{fig:mesh}. Meanwhile, compared with NeRF-based methods, FGO-SLAM shows better hole filling ability. This is attributed to our globally consistent opacity field, which bypasses TSDF fusion and directly extracts meshes from Gaussian primitives.

\subsection{Real-time performance Analysis}

In SLAM systems, real-time performance is also essential. We evaluated the average tracking and mapping time per frame, as well as the overall FPS, in the Office0 scene of the Replica dataset \cite{replica}. TABLE \ref{tab:time} compares our method with the baselines, demonstrating that our approach achieved the best results. This advantage is due to using only high-quality keyframes for mapping, which reduces data I/O time, achieving faster processing and meeting low-latency requirements.

\begin{figure}[!tbp]
	\centering
	\begin{subfigure}[b]{0.48\columnwidth}
		\centering
		\includegraphics[width=\textwidth]{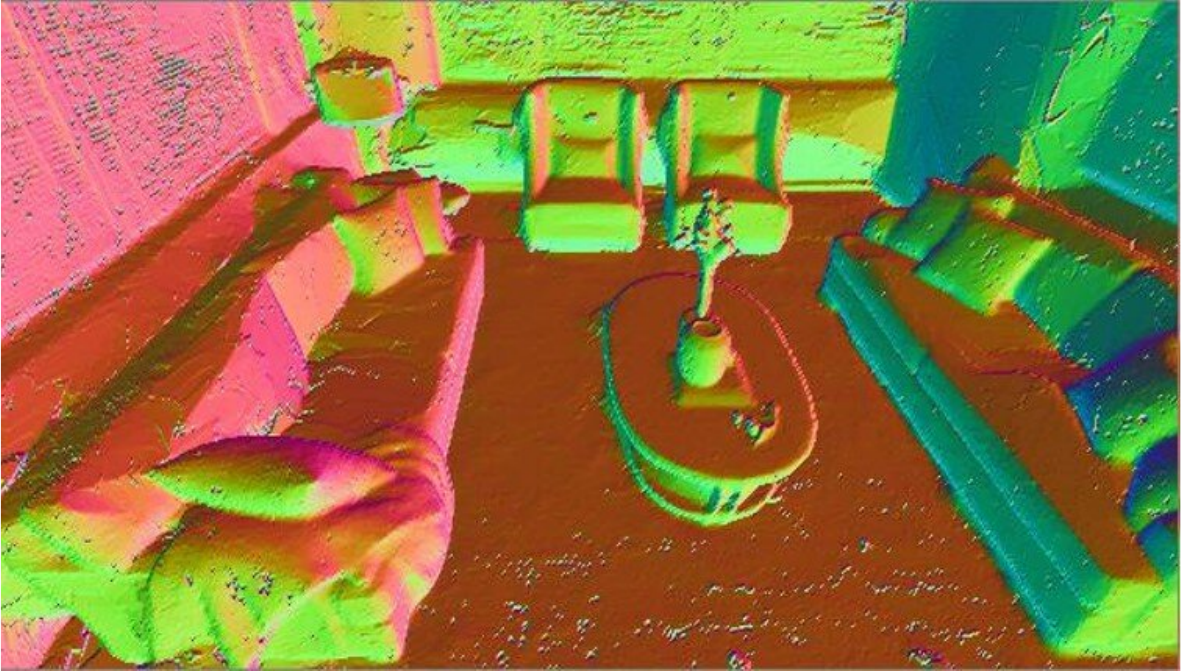}
		\caption{w/o regularization}
	\end{subfigure}
	\hfill
	\begin{subfigure}[b]{0.48\columnwidth}
		\centering
		\includegraphics[width=\textwidth]{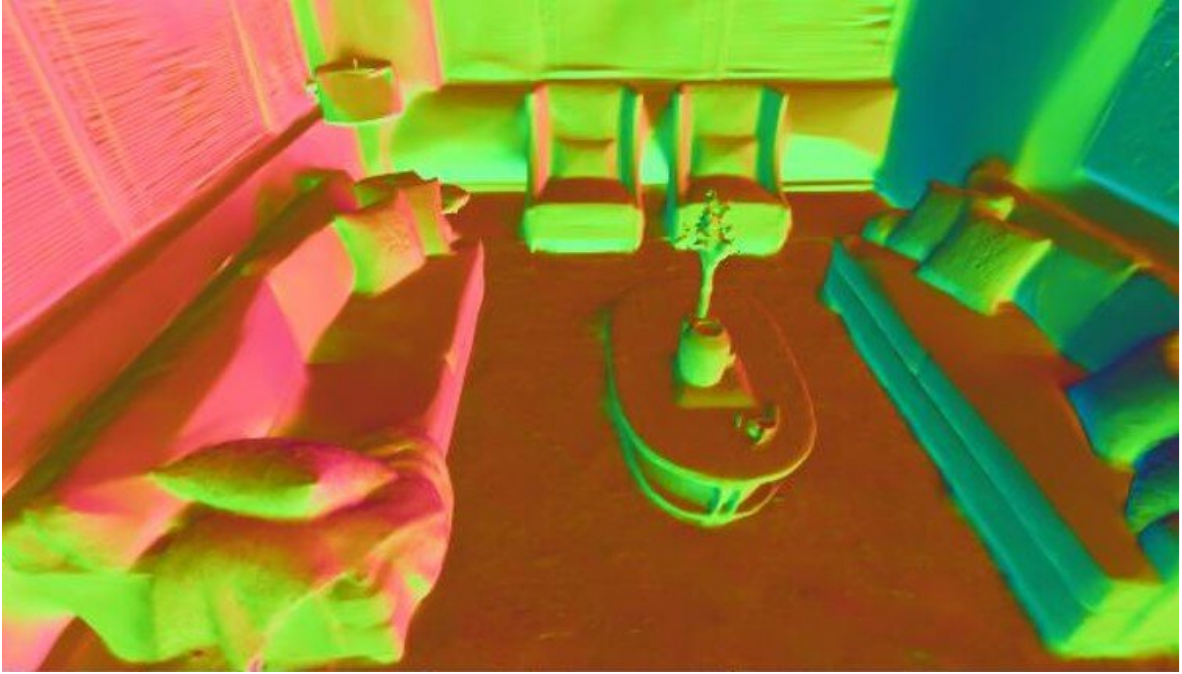}
		\caption{w regularization}
	\end{subfigure}
	\caption{\textbf{Ablation of the regularization.}}
	\label{fig:normal}
\end{figure}

\subsection{Ablation Study} 
In this section, we demonstrate the importance of the two key components, global adjustment and regularization terms, in our system.

\subsubsection{Global Adjustment} 

We assessed the impact of global adjustment on localization and mapping performance, with TABLE \ref{fig:ga} showing our quantitative results on the TUM RGB-D dataset. The use of global adjustment significantly improved trajectory prediction accuracy and rendered image quality. This shows that our approach enables rapid loop closing and effectively eliminates most accumulated errors without retraining the model.

\begin{table}[!tbp]
	\centering
	\caption{\textbf{Comparison of time performance on Office0 of the Replica dataset \cite{replica}.} The average tracking and mapping time per frame [s] ($\downarrow$), along with their FPS ($\uparrow$), were calculated.}
	\begin{tabular}{lccc}
		\toprule
		& Tracking /Frame $\downarrow$ & Mapping /Frame $\downarrow$ & FPS $\uparrow$\\ 
		\midrule
		NICE-SLAM \cite{Nice-slam}  & 1.07               & 1.04             & 0.85        \\
		ESLAM \cite{Eslam}      & 0.18               & 0.12             & 6.35        \\
		SplaTAM \cite{Splatam}    & 1.22               & 2.02             & 0.55        \\
		\textbf{Ours}       & \textbf{0.02}      & \textbf{0.09}    & \textbf{7.83} \\ 
		\bottomrule
	\end{tabular}
	\label{tab:time}
\end{table}

\begin{table}[!tbp]
	\caption{\textbf{Quantitative comparison of the ablation study on Global Adjustment.} ATE-RMSE [cm] ($\downarrow$) and PSNR [dB] ($\uparrow$) are used as the metrics, conducted on scene 0000\_00 of the ScanNet dataset \cite{Scannet}.}
	\centering
	\begin{tabular}{lcc}
		\toprule
		\textbf{} & ATE-RMSE $\downarrow$ & PSNR $\uparrow$ \\ 
		\midrule
		Ours w/o GA       & 7.59        & 23.54   \\
		Ours w GA    & \textbf{6.74}     & \textbf{25.12} \\ 
		\bottomrule
	\end{tabular}
\label{fig:ga}
\end{table}

\subsubsection{Regularization}

We assessed the impact of the regularization term on the system using the Replica dataset. Fig. \ref{fig:normal} compares surface extraction results with and without the regularization term. The inclusion of the regularization term results in a smoother overall reconstruction, particularly in floor areas.

\section{CONCLUSIONS}

We introduced FGO-SLAM, a Gaussian SLAM system using opacity radiance fields. We achieved globally consistent pose estimation through global adjustment. The use of opacity radiance fields for scene representation enabled us to achieve higher quality scenes and resolve the challenges of geometry extraction in Gaussian SLAM systems. Furthermore, FGO-SLAM is support for monocular cameras broadens its potential applications. We believe our work significantly contributes to advancing embodied intelligence. Future research will focus on overcoming challenging scenes, as well as improving reconstructed scenarios to make them easier to use for downstream tasks.

\section*{ACKNOWLEDGMENT}
This work was supported by the Youth Innovation Promotion Association of the CAS under Grant 2021115.



\bibliographystyle{IEEEtrans}

\bibliography{reference}

\end{document}